\documentclass[10pt, a4paper]{article}

\usepackage{lrec-coling2024} % this is the new style

\usepackage{inconsolata}
\usepackage{graphicx}
\usepackage{float}

\usepackage{amsmath}
\usepackage{amssymb}
\usepackage{hyperref}
\usepackage{caption}
\usepackage{subcaption}

\usepackage{natbib}
\usepackage[nameinlink]{cleveref}

\title{On the Scaling Laws of Geographical Representation\\ in Language Models}

\name{
    Nathan Godey$\mkern6mu^{1,2}$ \quad Éric de la Clergerie$^1$ \quad Benoît Sagot$^1$ \\
}

\address{$^1$Inria, $^2$Sorbonne Université \\
         Paris, France\\\{nathan.godey,eric.de\_la\_clergerie,benoit.sagot\}@inria.fr\\}

\abstract{
Language models have long been shown to embed geographical information in their hidden representations. This line of work has recently been revisited by extending this result to Large Language Models (LLMs). In this paper, we propose to fill the gap between well-established and recent literature by observing how geographical knowledge evolves when scaling language models. We show that geographical knowledge is observable even for tiny models, and that it scales consistently as we increase the model size. Notably, we observe that larger language models cannot mitigate the geographical bias that is inherent to the training data. 
 \\ \newline  }

\begin{document}

\maketitleabstract

\section{Introduction \& Related work}

In recent years, numerous studies analyzing the hidden representations of self-supervised language models have provided insights into how these models incorporate linguistic knowledge from their training data \citep{gupta-etal-2015-distributional,kohn-2015-whats,shi-etal-2016-string,ijcai2018p796,conneau-etal-2018-cram,jawahar-etal-2019-bert}. 

This line of work has been called probing, as most approaches are generally based on the training of classifiers---or \textit{probes}---upon frozen hidden representations.

Analyzing the representations of language models can point out sociocultural biases that were inherently learned by the models during training \citep{zhao-etal-2018-gender}, and training probes can help with mitigating these biases \citep{ravfogel-etal-2020-null, iskander-etal-2023-shielded}.

Among probing tasks, several works have focused on geographical representations that are implicitly embedded in language models. \citet{lotr} show that coordinates of places in the Middle-Earth can be predicted by just using the co-occurence matrix extracted from the Lord of the Rings novels. \citet{faisal-anastasopoulos-22-geographic} build networks from geographical representations based on monolingual and multilingual models of different sizes. They show that all models embed more accurate geographical representations for countries of the Global North.

This geographical discrepancy can be explained by biases that are inherent to the datasets used for pretraining \citet{faisal-etal-2022-dataset}. Imbalanced frequency distributions of geographical references in pretraining data causes distortions in the representational space \citep{zhou2021frequencybased}. These distortions lead to a loss in the models' ability to differentiate between under-represented locations.

Recently, \citet{gurnee2023language} have probed large language models from the Llama-2 suite \citep{touvron2023llama} to extract coordinates of prompted locations from hidden representations across layers. They show that models ranging from 7B to 70B parameters are able to convincingly embed geographical coordinates on a world map when representing basic prompts.

In this work, we propose to extend the analysis by \citet{gurnee2023language} to smaller language models, in order to observe how scale affects the ability of models to implicitly embed geographical information from raw training data. We show that such ability consistently improves with model size, and that even tiny models are able to produce visually meaningful world maps.

We make several contributions:
\begin{itemize}
    \item We show that geographical information can be extracted to a certain extent from representations at every model scale;
    \item We observe that larger models are more geographically biased than their smaller counterparts;
    \item We find that the performance of models in terms of geographical probing is correlated with the frequency of corresponding country names in the training data.
\end{itemize} 

\section{Scaling Laws of Geographical Probing}
\label{sec:scaling}
\begin{figure*}[h]
    \centering
    \begin{subfigure}[b]{0.43\textwidth}
         \includegraphics[trim={0 0 0 0.7cm},clip,width=\linewidth]{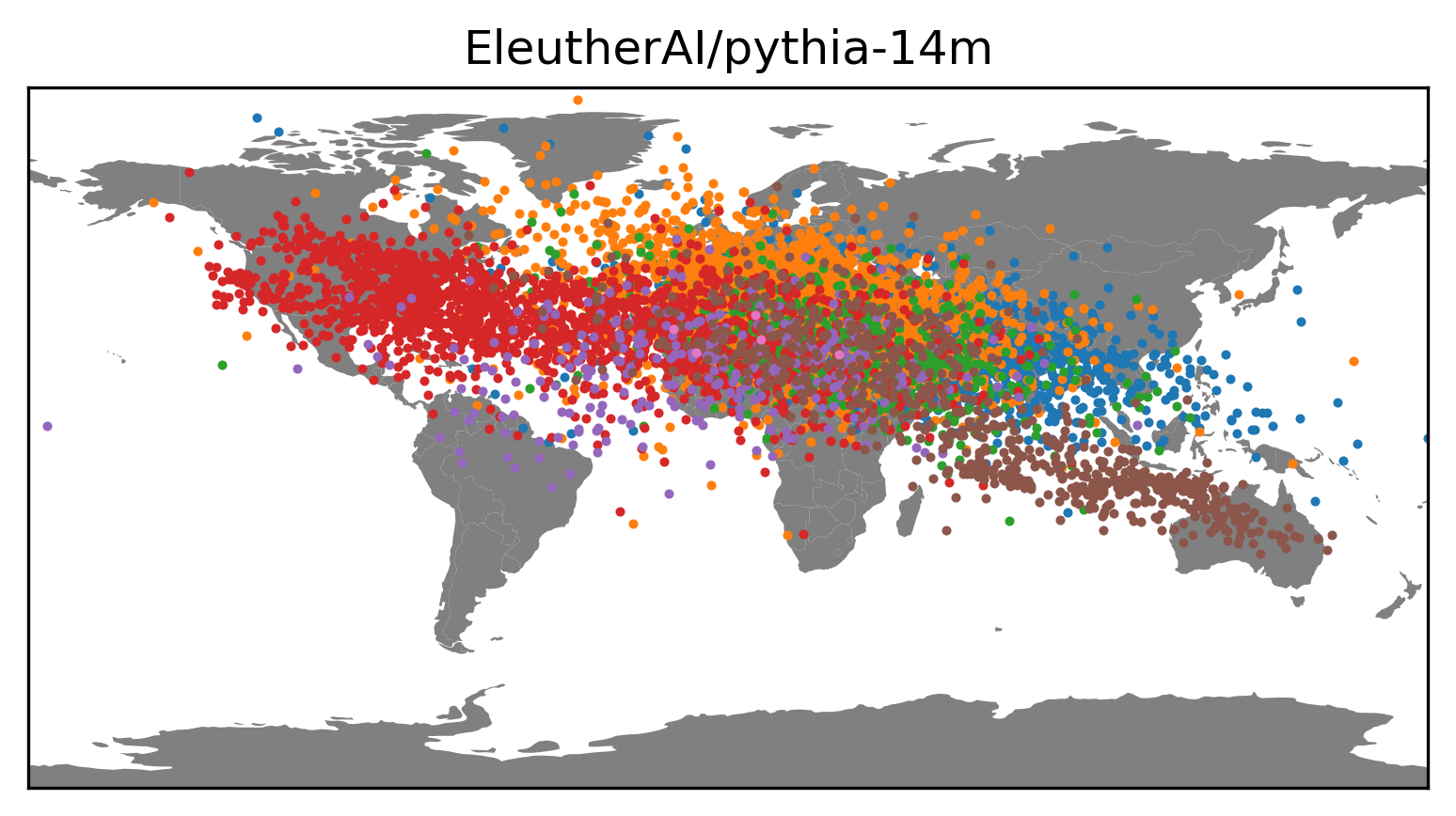}
         \caption{Pythia 14M ($R^2 = 34.34$)}
         \label{fig:14m_map}
         \vspace{1em}
    \end{subfigure}
    \begin{subfigure}[b]{0.43\textwidth}
         \includegraphics[trim={0 0 0 0.7cm},clip,width=\linewidth]{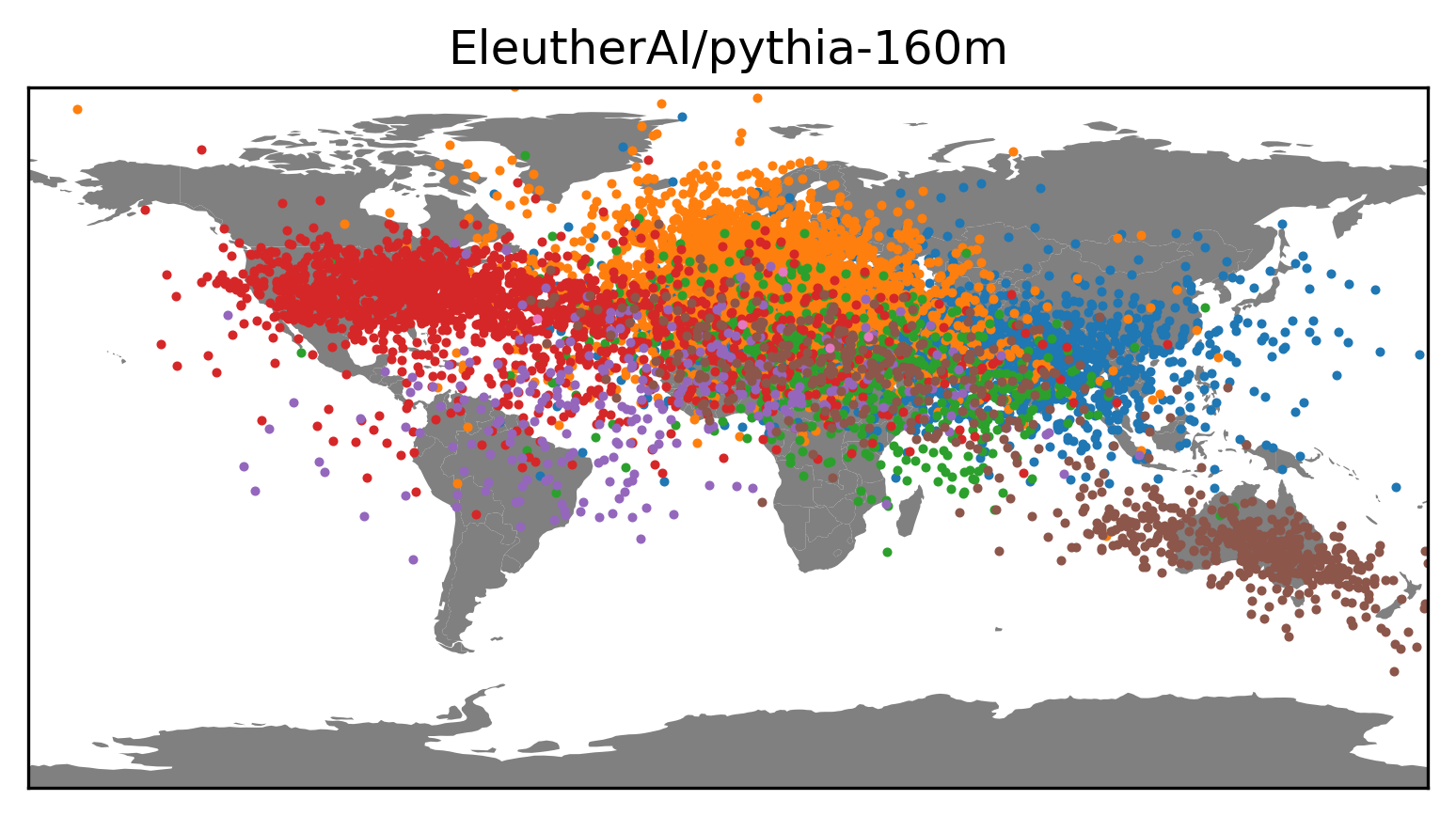}
         \caption{Pythia 160M ($R^2 = 55.28$)}
         \label{fig:160m_map}
        \vspace{1em}
    \end{subfigure}
    \begin{subfigure}[b]{0.43\textwidth}
         \includegraphics[trim={0 0 0 0.7cm},clip,width=\linewidth]{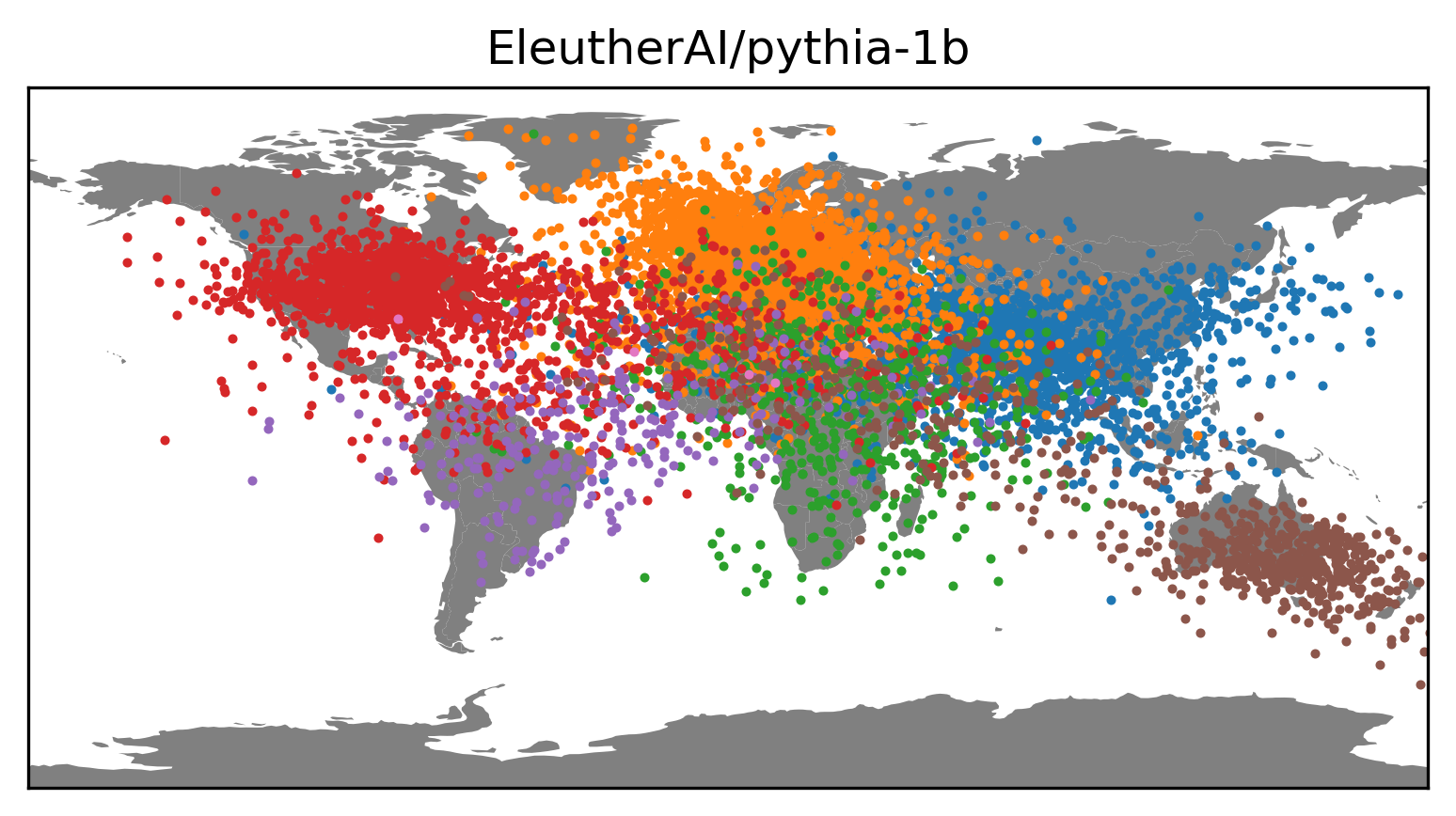}
         \caption{Pythia 1B ($R^2 = 67.94$)}
         \label{fig:1b_map}
    \end{subfigure}
    \begin{subfigure}[b]{0.43\textwidth}
         \includegraphics[trim={0 0 0 0.7cm},clip,width=\linewidth]{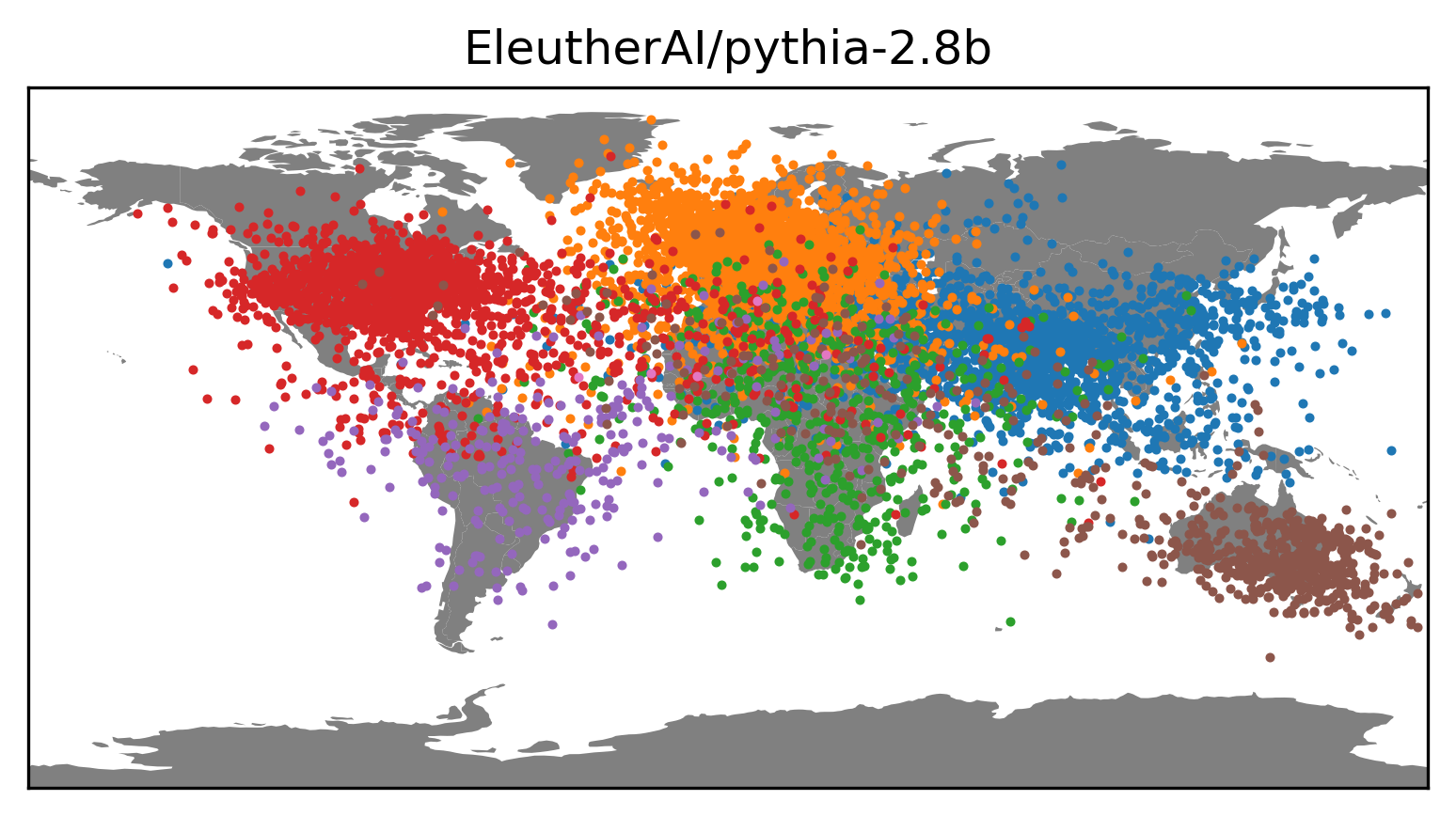}
         \caption{Pythia 2.8B ($R^2 = 74.97$)}
         \label{fig:2.8b_map}
    \end{subfigure}
    \caption{Predicted coordinates of test set instances for different model sizes. Each color represents a different continent.}
    \label{fig:maps}
\end{figure*}

In this section, we train geographical probes for a wide variety of models at different scales.

\subsection{Methodology}

We use the World dataset from \citet{gurnee2023language} as a geographical data source. It contains 39,504 location names from the whole world along with corresponding longitude and latitude. We use the same train-test split strategy as in the original article, thus keeping 20\% of samples for testing purposes.

For each location name $X$, we prompt models with the text: ``\textit{Where is $X$ in the world?}''. We then infer with a given model on the whole dataset, and use the last token belonging to the entity $X$ as the model's representation. To follow the linear probing paradigm used in \citet{gurnee2023language}, we train a Ridge linear regressor \citep{ridge} to predict latitude and longitude based on the model's representations. We then measure the probe's performance on the test set using the $R^2$ correlation coefficient.

\subsection{Results}
In \autoref{fig:maps}, we display the predictions of the probe for the most performant layer, which is generally the last one. We observe that geographical information can be extracted from models even for a very small parameter count. The performance of the probes seem to increase with the model size.

\begin{figure}[h]
    \centering
    \begin{subfigure}[b]{0.37\textwidth}
         \includegraphics[width=\linewidth]{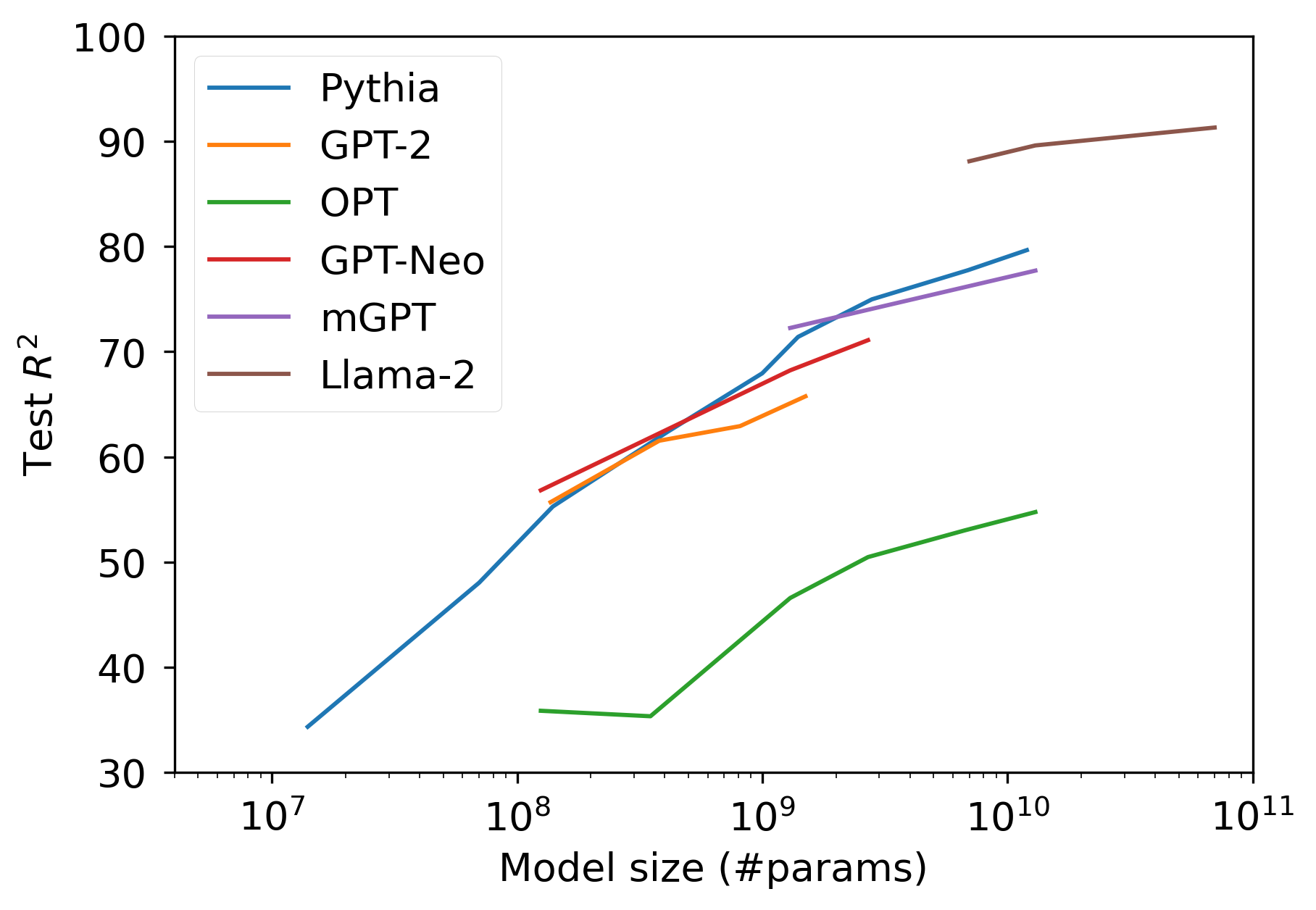}
         \caption{Decoder models}
         \label{fig:decoder_evol}
         \vspace{1em}
    \end{subfigure}
    \begin{subfigure}[b]{0.37\textwidth}
         \includegraphics[width=\linewidth]{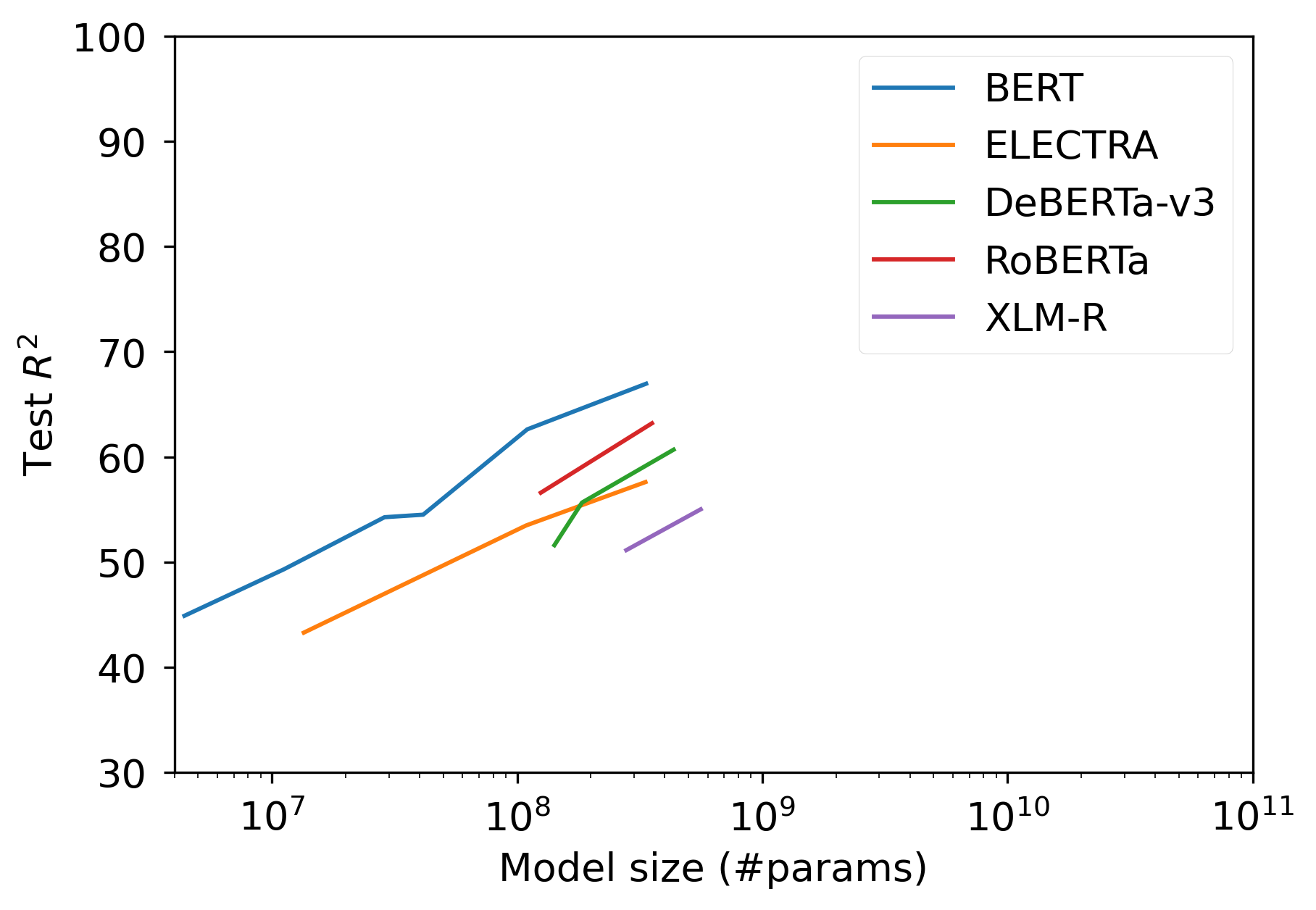}
         \caption{Encoder models}
         \label{fig:encoder_evol}
    \end{subfigure}
    \caption{Evolution of the $R^2$ coefficient on the test set for various model suites.}
    \label{fig:evol}
\end{figure}

We show in \autoref{fig:evol} that the performance of language models evolves consistently with model size, regardless of the architecture. We validate this property on several decoder model families: GPT-2 \citep{gpt2}, OPT \citep{zhang2022opt}, Pythia \citep{pythia}, GPT-Neo \citep{gpt-neo}, the multilingual mGPT \citep{shliazhko2023mgpt}, and Llama-2 \citep{touvron2023llama}. We also display results for several encoder models: BERT \citep{devlin-etal-2019-bert,turc2019wellread}, RoBERTa \citep{roberta}, ELECTRA \citep{electra}, and DeBERTa-v3 \citep{deberta}. This property also applies for encoder models, for which we notice that the BERT suite unexpectedly outperforms its counterparts. The performance of encoder models is comparable with the one of equivalent decoder models. We can underline the fact that BERT-Large (336M parameters) is as accurate as the three times larger Pythia-1B.

Interestingly, the multilingual XLM-R \citep{conneau2020unsupervised} underperforms its counterparts, even though multilingual data must have increased the training data's geographical diversity to some extent \citep{faisal-anastasopoulos-2021-investigating}. The mGPT suite also slightly underperforms Pythia models at equivalent model sizes.

We verified that the better performance of larger models was not solely related with the ability of the probes to extract better patterns from their higher-dimensionality hidden representations. We achieved this by concatenating representations with themselves to increase dimensionality without introducing novel knowledge. It led to slightly worse performance for all tested models, thus showing that performance was not a consequence of dimensionality alone.

\section{Geographical Bias and Scale}

In \autoref{fig:maps}, it seems at first glance that as the model size increases, the predictions tend to be more accurate for locations of the Southern Hemisphere. In this section, we propose to quantify this hypothesized behavior by measuring the bias across countries and continents for various scales. We also correlate the models' accuracy with both lexical and geographical factors.

\subsection{Measuring bias}

\begin{figure}
    \centering
    \includegraphics[width=0.9\linewidth]{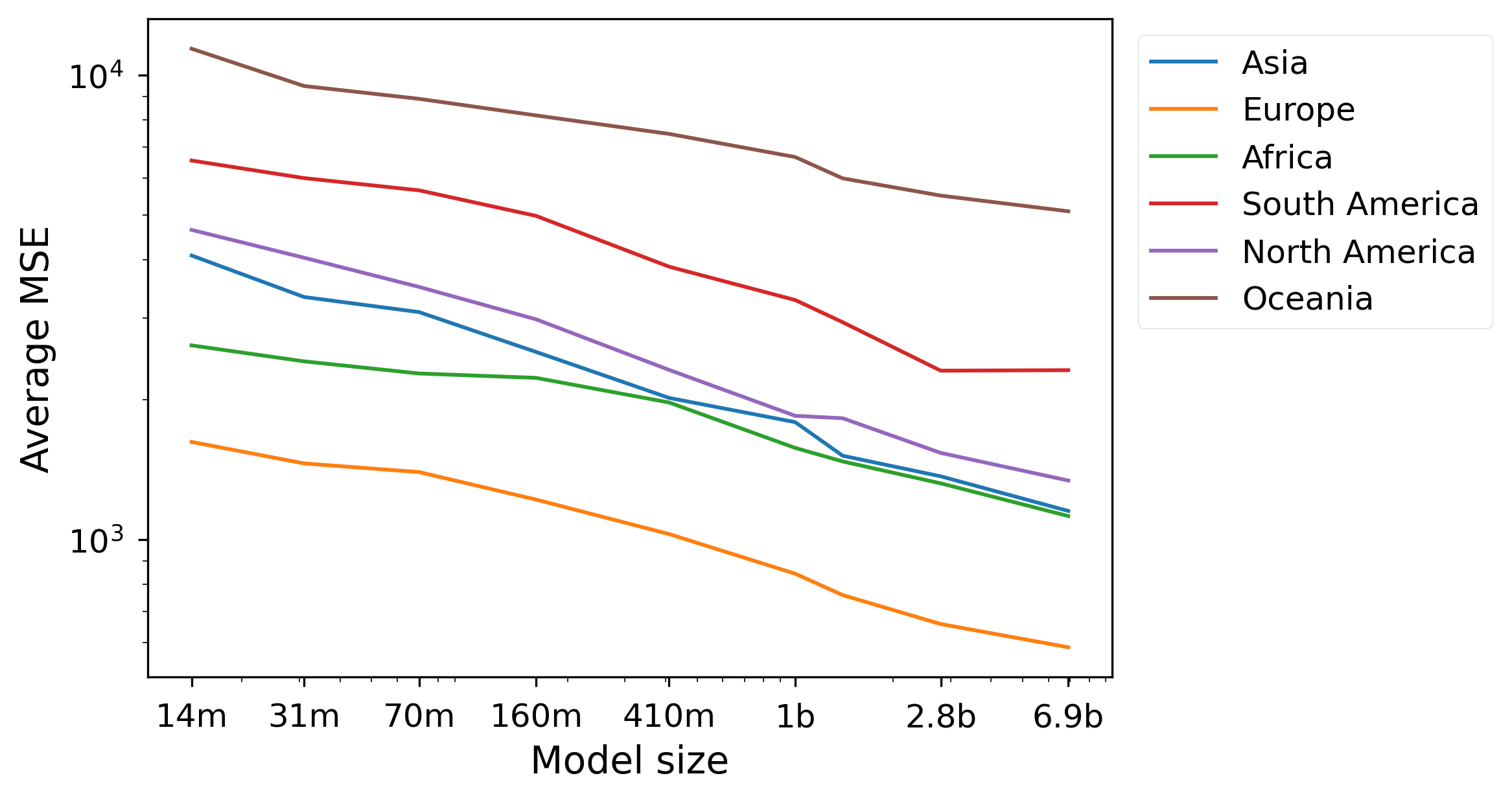}
    \caption{Average MSE by continent for different sizes in the Pythia suite.}
    \label{fig:continent_perf}
\end{figure}

We group probe performance as measured by mean-squared error (MSE) on predicted coordinates, and average measures by continent in \autoref{fig:continent_perf}. While we notice that the performance increases consistently for every continent, we do not observe a significant reduction in the performance gap across continents as model size increases.

To measure the heterogeneity of the probing performance of language models across countries, we use the Gini coefficient \citep{gini1912variabilita} that is widely used in economics. Given a series of observed variables $(x_i)_{i\in[1, N]}$, the Gini coefficient is defined as:
$$
Gini(x) = \frac{\sum_{i,j \in [1, N]} |x_i - x_j|}{N \cdot \sum_{i = 1}^{N} x_i}
$$

A Gini coefficient of 1 reflects perfect heterogeneity, while a Gini of 0 implies perfect homogeneity.

\begin{figure}
    \centering
    \includegraphics[width=0.75\linewidth]{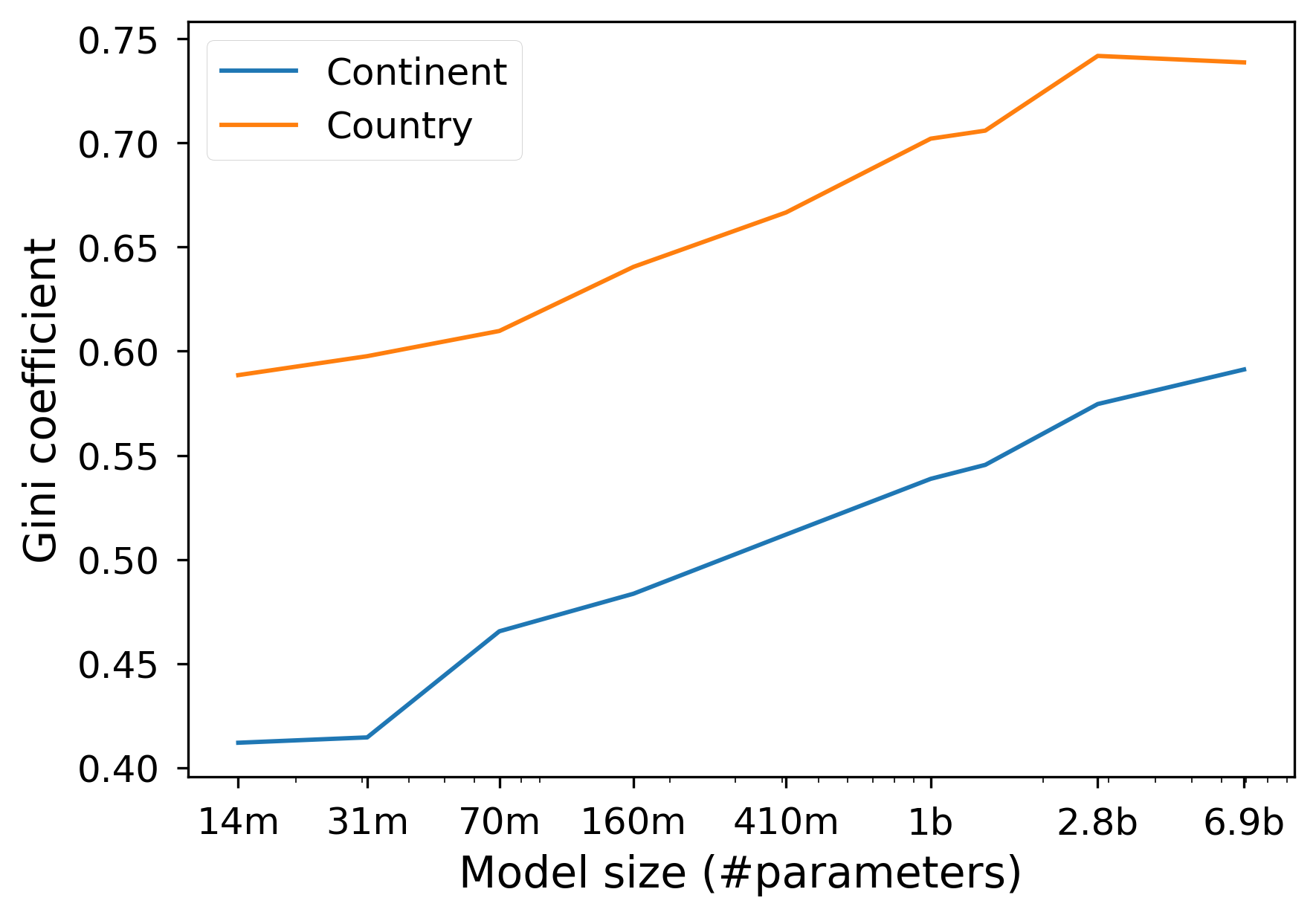}
    \caption{Gini coefficients of MSE on the test set averaged by country or by continent, as model size increases.}
    \label{fig:ginis}
\end{figure}

\autoref{fig:ginis} shows that the larger the model is, the more heterogeneous the probe performance is across countries and continents. This contradicts the impression given by \autoref{fig:maps}, and shows that scale does not solve the geographical discrepancy caused by bias inherent to the training data.

\begin{figure}
    \centering
    \includegraphics[width=0.9\linewidth]{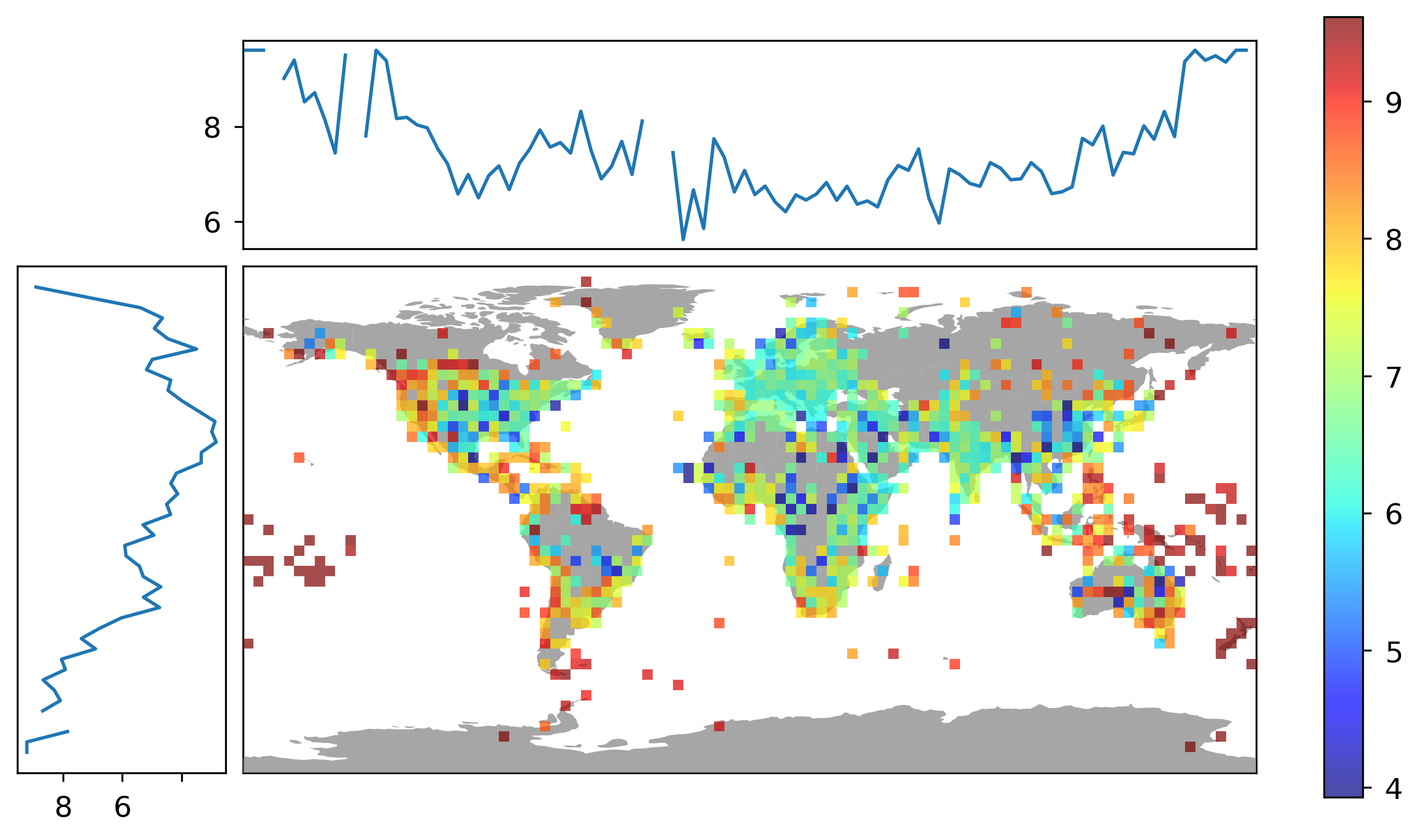}
    \caption{Test log-MSE for Pythia-1B as plotted on a World map.}
    \label{fig:heatmap}
\end{figure}

In \autoref{fig:heatmap}, we locally average log-MSE on a World map, and report results agglomerated accorded to latitude and longitude. We clearly observe that the model performs poorly in Oceania, South Asia and South America. We also see that the error is minimal around the latitude of North America and Europe, while it increases in the Southern Hemisphere.

\subsection{Identifying sources of bias}
We attempt to correlate the performance of our geographical probes with several factors. First, the dataset from \citep{gurnee2023language} provides each location with an estimate of the corresponding population count when relevant. We also consider training data distribution as a potential factor of heterogeneity. Finally, we consider latitude and longitude as potential factors of bias.

To account for training data distribution, we look for exact string matches of country names from the \citet{gurnee2023language} dataset in an extract of The Pile \citeplanguageresource{gao2020pile} containing 3.5 million samples \footnote{\url{https://huggingface.co/datasets/ola13/small-the\_pile}}. We select this dataset as it was used to pretrain the models from the Pythia suite \citep{pythia} we evaluate in this section. We find 15 million matches, covering 98\% of the countries of the dataset.

We do not count occurrences of location names directly, as matching locations on the basis of their names does not account for named entity ambiguity. An example of ambiguous location name is \textit{Fully}, which is a town in Switzerland. An exact match strategy overestimates by large margins the occurrence count of this location, because of the corresponding English word \textit{fully}. Disambiguation techniques have been designed \citep{hoffart-etal-2011-robust, orr2020bootleg}, but we prefer to avoid the risk of bias propagation and the cost of using such methods on a large corpus.

\begin{figure}
    \centering
    \includegraphics[width=0.9\linewidth]{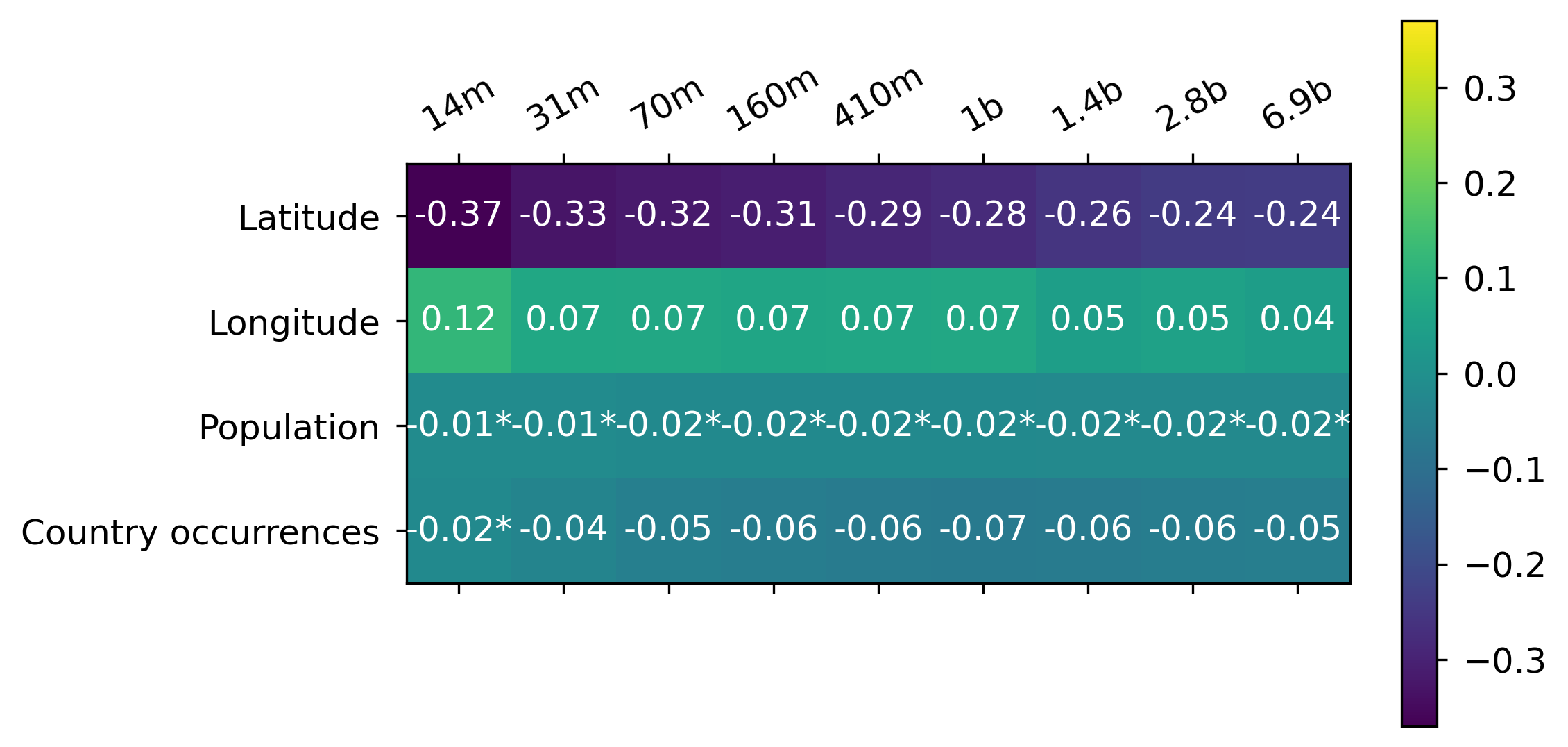}
    \caption{Pearson correlation coefficients of various factors with location-wise MSE, for several Pythia model sizes. *: Tests that yielded p-values above 0.05.}
    \label{fig:correl}
\end{figure}

We display Pearson correlations between each of the aforementioned factors and the entity-level MSE for each model size in \autoref{fig:correl}. As in \autoref{fig:14m_map}, we observe that the error on coordinates prediction is negatively correlated with the latitude, i.e. southern locations are less accurately identified. This correlation slowly decays as the model size increases. Meanwhile, longitude seems to be mildly correlated with the probe performance.

Interestingly, the population count is not correlated with the error level. The occurrence count of the location country is negatively correlated with the error level, thus showing that the more country names appear in the training dataset, the more the probes are able to recover coordinates from locations in these countries. However, this correlation is mild and even below the significance threshold for the smallest model.

We also measure the correlation between country occurrences and other metrics to account for the bias inherent to the data. We observe that country name occurrences are positively correlated with latitude with a p-value of 0.06, and not correlated with the longitude. More importantly, the population count of a country and the count of this country name in the data are heavily correlated (factor of +0.52 and p-value of 3e-23). Thus, even though the data seems guided by demographic factors, this is not the case of the model's representations.

\section{Discussion}
\label{sec:discussion}
We believe that quantifying sociocultural bias in representations of language models and pretraining datasets allows to better understand the roots of the biases that can be observed during generation.

\citet{parrots_bender} discuss the relevance of scaling models to ever larger magnitudes, with regard to environmental and financial costs. Our study shows that scale can also increase language modeling bias when it comes to geographical representation, given a pretraining dataset. We advocate in favor of measuring and mitigating bias in pretraining datasets to avoid scaling bias along with performance.

\section*{Conclusion}
In this study, we show that a wide variety of language models, varying in architecture and sizes, implicitly embed geographical data to some extent. As we consider larger models, the performance of geographical probes consistently increases towards levels shown in \citet{gurnee2023language}.

We show numerically that the geographical probe performance is correlated with latitude across model sizes, but also with the number of occurrence of corresponding country names in the pretraining data. Conversely, the population count of the location seems uncorrelated with the probe performance. This indicates that a minority of people benefit from better geographical understanding when using language models, which does not maximize the social utility of these systems.

While it may initially seem that this performance increase mitigates heterogeneity between Southern and Northern countries, we actually show that larger models tend to be more biased according to the Gini coefficient taken on prediction error. This tends to show that scaling language models can amplify discrepancies in their geographical knowledge.

\section*{Acknowledgements}
This work was funded by the last authors' chair in the PRAIRIE institute, funded by the French national agency ANR as part of the ``Investissements d'avenir'' program under the reference ANR-19-P3IA-0001.
We thank Stella Biderman for her insightful advice.

% We would like to thank Roman Castagné for useful discussions that led to focusing on observing the effect of anisotropy in the self-attention process.

\section*{Bibliographical References}
\label{sec:reference}
\bibliography{anthology,custom}
\bibliographystyle{lrec-coling2024-natbib}

\section*{Language Resource References}
\label{lr:ref}
\bibliographylanguageresource{lrec_lr_natbib}
\bibliographystylelanguageresource{lrec-coling2024-natbib}

\clearpage

\appendix

\end{document}